\documentclass{article}

\usepackage{microtype}
\usepackage{graphicx}
\usepackage{subcaption}
\usepackage{booktabs} 
\usepackage{xcolor}
\usepackage{amsmath}
\usepackage{amssymb}

\usepackage{hyperref}

\usepackage[accepted]{icml2020}

\usepackage{selectp}

\DeclareMathOperator*{\argmax}{arg\,max}

\def\*#1{\mathbf{#1}}

\begin{document}

\twocolumn[
  \icmltitle{Uncertainty Estimation Using a Single Deep Deterministic Neural Network}



  \icmlsetsymbol{equal}{*}

  \begin{icmlauthorlist}
    \icmlauthor{Joost van Amersfoort}{cs}
    \icmlauthor{Lewis Smith}{cs}
    \icmlauthor{Yee Whye Teh}{stats}
    \icmlauthor{Yarin Gal}{cs}
  \end{icmlauthorlist}

  \icmlaffiliation{cs}{OATML, Department of Computer Science, University of Oxford}
  \icmlaffiliation{stats}{Department of Statistics, University of Oxford}

  \icmlcorrespondingauthor{Joost van Amersfoort}{joost.van.amersfoort@cs.ox.ac.uk}

  \icmlkeywords{Uncertainty Estimation, Deep Learning}

  \vskip 0.3in]



\printAffiliationsAndNotice{}  

\begin{abstract}
  We propose a method for training a deterministic deep model that can find and
  reject out of distribution data points at test time with a single forward
  pass. Our approach, deterministic uncertainty quantification (DUQ), builds
  upon ideas of RBF networks. We scale training in these with a novel loss
  function and centroid updating scheme and match the accuracy of softmax
  models. By enforcing detectability of changes in the input using a gradient
  penalty, we are able to reliably detect out of distribution data. Our
  uncertainty quantification scales well to large datasets, and using a single
  model, we improve upon or match Deep Ensembles in out of distribution
  detection on notable difficult dataset pairs such as FashionMNIST vs. MNIST,
  and CIFAR-10 vs. SVHN.
\end{abstract}

\section{Introduction}

\begin{figure}[t]
  \begin{subfigure}{0.5\linewidth}
    \centering
    \includegraphics[width=\linewidth]{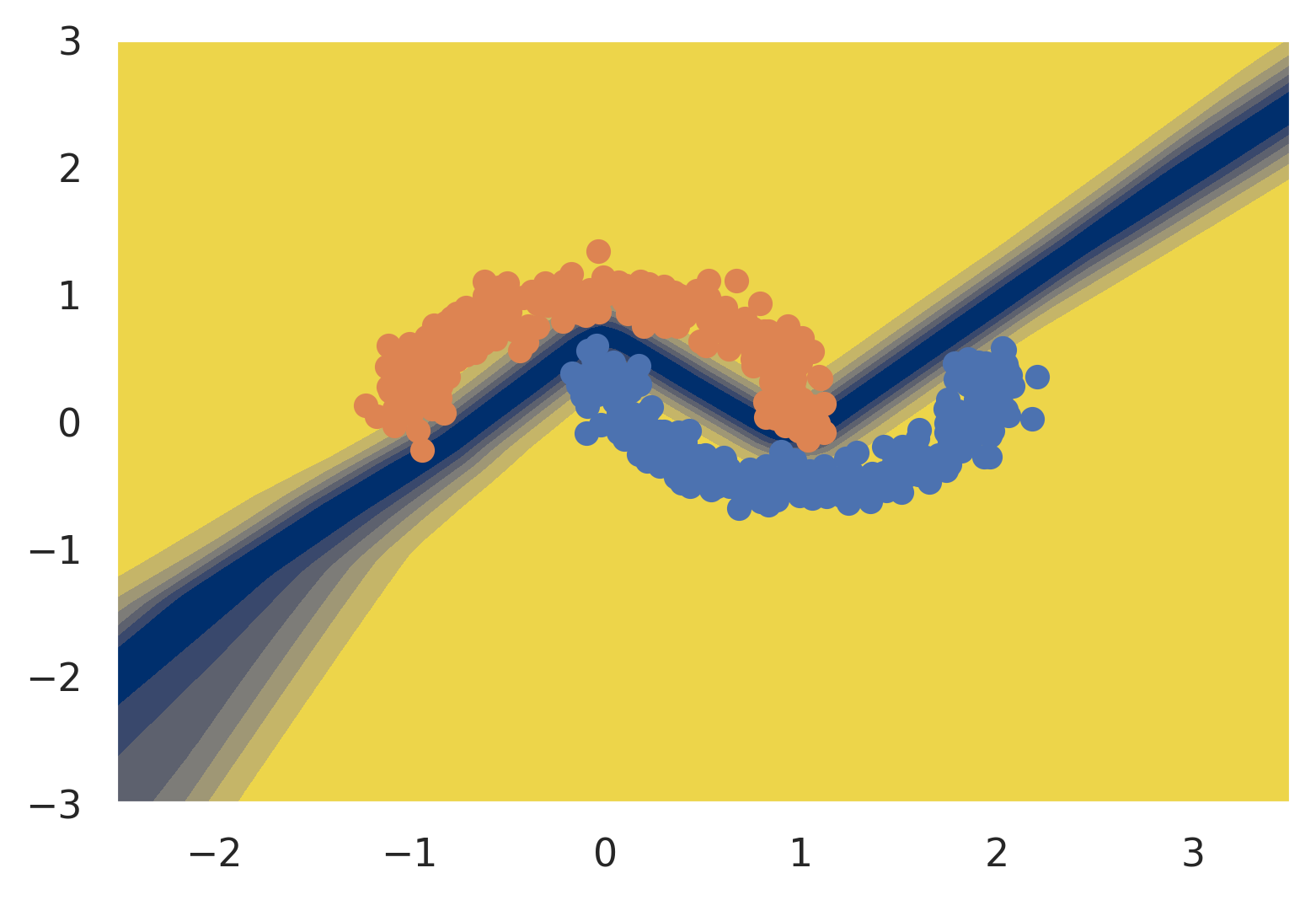}
    \caption{Deep Ensembles}
    \label{two_moons_deep_ensembles}
  \end{subfigure}%
  \begin{subfigure}{0.5\linewidth}
    \centering
    \includegraphics[width=\linewidth]{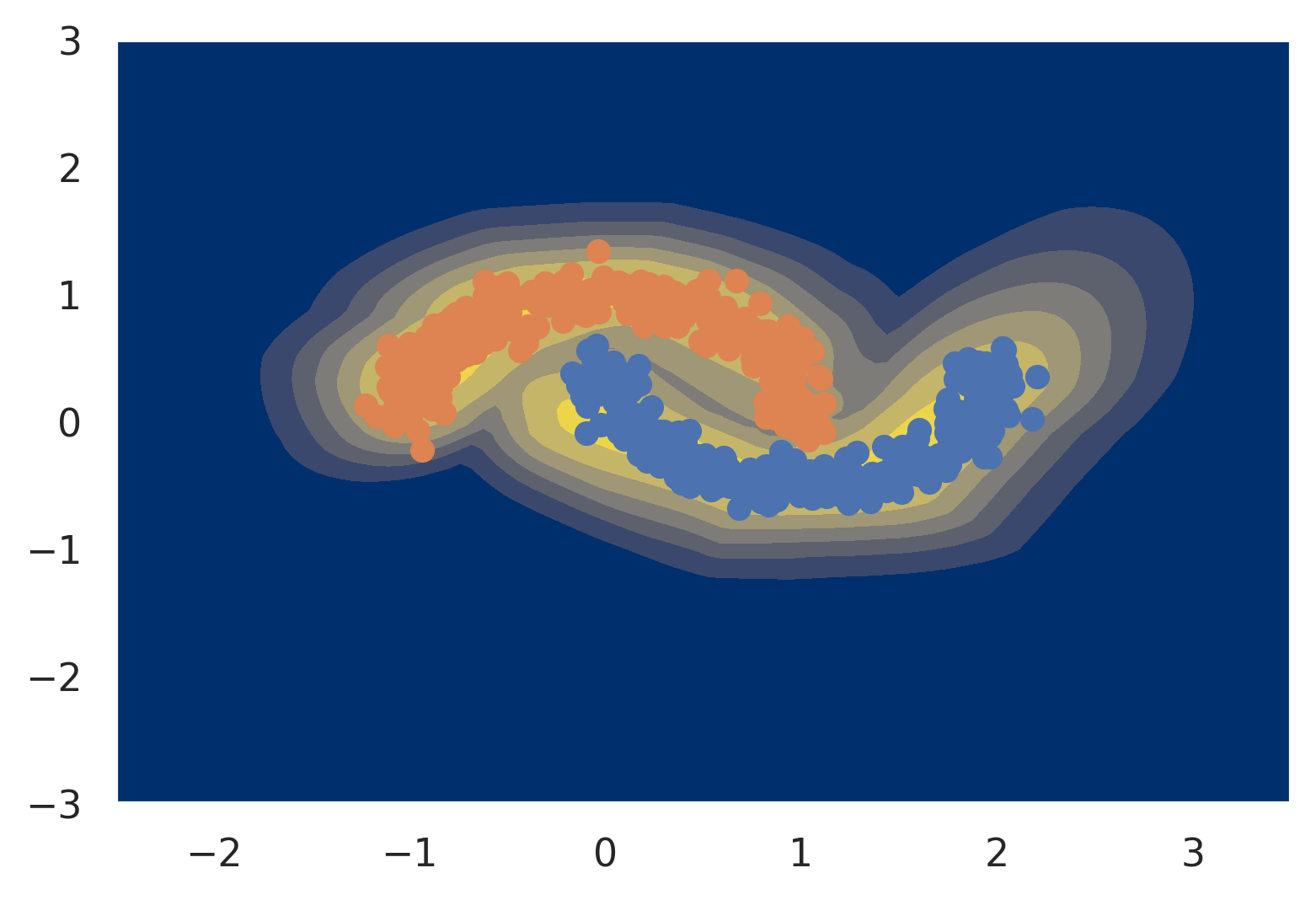}
    \caption{Our model - DUQ}
    \label{two_moons_duq}
  \end{subfigure}
  \caption{Uncertainty results on two moons dataset. Yellow indicates high
    certainty, while blue indicates uncertainty. DUQ is certain only on the data
    distribution, and uncertain away from it: the ideal result. Deep Ensembles is
    uncertain only along the decision boundary, and certain elsewhere.}
  \label{two_moons_new}
\end{figure}

Estimating uncertainty reliably and efficiently has remained an open problem
with many important applications such as guiding exploration in Reinforcement
Learning \citep{osband2016deep} or as a method for selecting data points for
which to acquire labels in Active Learning \citep{houlsby2011bayesian}. Until
now, most approaches for estimating uncertainty in deep learning rely on
ensembling \citep{lakshminarayanan2017simple} or Monte Carlo sampling
\citep{gal2016dropout}. In this paper, we introduce a deep model that is able to
estimate uncertainty in a single forward pass. We call our model DUQ,
Deterministic Uncertainty Quantification, and we construct it by re-examining
ideas originally suggested in the 90s. We combine these with recent advances and
make a number of improvements which enable scalable training of modern deep
learning architectures. We evaluate our model against the current best approach
for estimating uncertainty in Deep Learning, Deep Ensembles, and show that DUQ
compares favourably on a number of evaluations, such as out of distribution
(OoD) detection of FashionMNIST vs MNIST, and CIFAR vs. SVHN. We visualise how
DUQ performs on the two moons dataset in Figure \ref{two_moons_new}. We see that
DUQ is only certain on the training data, and its certainty decreases away from
it. Deep Ensembles are not able to obtain meaningful uncertainty on this
dataset, because of a lack of diversity in the different models in the ensemble.
We make our code publicly available\footnote{\url{https://github.com/y0ast/deterministic-uncertainty-quantification}}.

DUQ consists of a deep model and a set of feature vectors corresponding to the
different classes (\textit{centroids}). A prediction is made by computing a
kernel function, a distance function, between the feature vector computed by the
model and the centroids. This type of model is called an \textit{RBF network}
\citep{lecun1998gradient} and uncertainty is measured as the distance between
the model output and the closest centroid. A data point for which the feature
vector is far away from all centroids does not belong to any class and can be
considered out of distribution. In this paper, we define uncertainty to be
\textit{predictive} uncertainty.

The model is trained by minimising the distance to the correct centroid, while
maximising it with respect to the others. This incentivises the model to put the
features of training data close to a particular centroid, however there is no
mechanism that dictates what should happen away from the training data.
Therefore we need to enforce that DUQ is sensitive to changes in the input, such
that we can reliably detect out of distribution data and avoid mapping out of
distribution data to in distribution feature representations --- an effect we call
\textit{feature collapse}. The upper bound of this sensitivity can be quantified
by the Lipschitz constant of the model. We are interested in models for which
this sensitivity is not too low, but also not too high, because that could hurt
generalisation and optimisation. DUQ achieves this result by regularising the
Jacobian with respect to the input, as was first introduced by
\citet{drucker1992improving}.

In practice, RBF networks prove difficult to optimise, because of instability of
the centroids and a saturating loss. We propose to make training stable by
updating the centroids using an exponential moving average of the feature
vectors of the data points assigned to them, as was introduced in
\citet{van2017neural}. We use a ``one vs the rest'' loss function minimising the
distance to the correct centroid, while maximising the other distances. We find
that these two changes stabilise training and lead to accuracies that are
similar to the standard softmax and cross entropy set up on standard datasets
such as FashionMNIST and CIFAR-10.

Uncertainty quantification in deep neural networks with a softmax output is
generally done by measuring the entropy of the predictive distribution, so
the maximally uncertain output is achieved by uniformly assigning probabilities
over all the classes. The only way to achieve a uniform output for out of
distribution data, is by training on additional data and hoping it generalises
to out of distribution samples at test time. This does not happen in practice,
and it is found that the only uncertainty that can reliably be captured by looking
at the entropy of the softmax distribution is aleatoric uncertainty
\citep{gal2016uncertainty, hein2019relu}. In DUQ, it is possible to predict that none of the
classes seen during training is a good fit, when the distance between the model
output and all centroids is large.

The contributions of this paper are as follows:
\begin{itemize}
  \item We stabilise training of RBF networks and show, for the first time, that
        these type of models can achieve competitive accuracy versus softmax models.
  \item We show how two-sided Jacobian regularisation makes it possible to obtain reliable uncertainty
        estimates for RBF networks.
  \item We obtain excellent uncertainty in a single forward pass,
        while maintaining competitive accuracy.
\end{itemize}
\begin{figure}[t]
  \centering
  \includegraphics[width=0.9\linewidth]{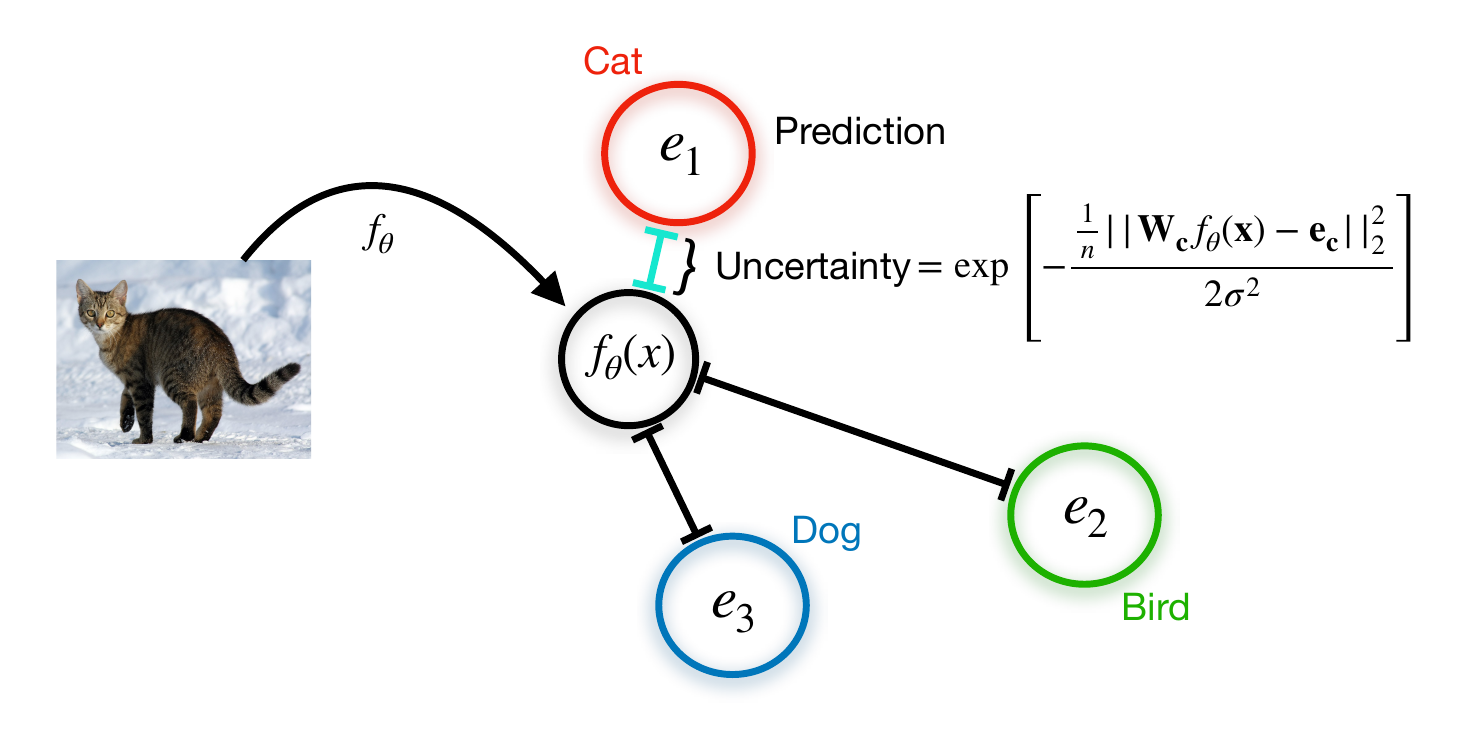}
  \caption{A depiction of the architecture of DUQ. The input is mapped to
    feature space, where it is assigned to the closest centroid. The distance to
    that centroid is the uncertainty.}
\end{figure}
\section{Methods}

DUQ consists of a deep feature extractor, such as a ResNet \citep{he2016deep},
but without the softmax layer. Instead, we have one learnable weight matrix
$W_c$ per class, $c$. Using the output and the class centroids, we compute the
exponentiated distance between the model output and the centroids:
\begin{align}
  K_c(f_\theta(\*x), \*e_c) = \exp \left [ -\frac{\frac1n||\*W_c f_\theta(\*x) - \*e_c||_2^2}{2\sigma^2} \right ],
\end{align}
with $f_\theta: \mathbb{R}^m \rightarrow \mathbb{R}^d$ our model, $m$ the input
dimension, $d$ the output dimension, and parameters $\theta$. $\*e_c$ is the
centroid for class $c$, a vector of length $n$. $\*W_c$ is a weight matrix of size
$n$ (centroid size) by $d$ (feature extractor output size) and $\sigma$ a hyper
parameter sometimes called the length scale. This function is also referred to
as a Radial Basis Function (RBF) kernel. The class dependent weight matrix
allows features insensitivity on a class by class basis, minimising the
potential for feature collapse. A prediction is made by taking the class $c$
with the maximum correlation (minimum distance) between data point $\*x$ and
class centroids $\*E = \{\*e_1, \dotsc, \*e_C\}$:
\begin{align}
  \label{decision_function}
  \argmax_c K_c(f_\theta(\*x), \*e_c).
\end{align}
we define the uncertainty in this model as the distance to the closest centroid,
i.e. replacing the $\argmax$ operator by a $\max$ in Equation
\eqref{decision_function}.

The loss function is the sum of the binary cross entropy between each class'
kernel value $K_c(\cdot, \*e_c)$, and a one-hot (binary) encoding of the label.
For a particular data point $\{\*x, \*y\}$ in our data set $\{X, Y\}$:
\begin{align}
  \label{loss}
  L(\*x, \*y) =
  - \sum_c y_{c} \log (K_c) + (1 - y_{c}) \log (1 - K_c)
\end{align}
where we shortened $K(f_\theta(\*x), \*e_c)$ as $K_c$. During training, we
average the loss over a minibatch of data points, and perform stochastic
gradient descent on $\theta$ and $\mathcal{W} = \{\*W_1, \cdots, \*W_c\}$. The
class centroids, $\*E$, are updated using an exponential moving average of the
feature vectors of data points belonging to that class. If the model parameters,
$\theta$ and $\mathcal{W}$, are held constant, then this update rule leads to the
closed form solution for the centroids that minimises the loss:
\begin{align}
  n_{c,t}    & = \gamma * n_{c, t-1} + (1 - \gamma) * n_{c,t}                              \\
  \*m_{c,t}  & = \gamma * \*m_{c, t-1} + (1 - \gamma) \sum_{i} \*W_c f_\theta(\*x_{c,t,i}) \\
  \*e_{c, t} & = \frac{\*m_{c,t}}{n_{c,t}}
\end{align}
where $n_{c,t}$ is the number of data points assigned to class $c$ in minibatch
$t$, $\*x_{c,t,i}$ is element $i$ of a minibatch at time $t$, with class $c$.
$\gamma$ is the momentum, which we usually set between [0.99, 0.999]. This
method of updating centroids was introduced in the Appendix of
\citet{van2017neural} for updating quantised latent variables. The high
momentum leads to stable optimisation that is robust to initialisation.

The proposed set up leads to the centroids being pushed further away at each
minibatch, without converging to a stable point. We avoid this by regularising
the $l_2$ norm of $\theta$. This restricts the model to sensible solutions and
aids optimisation.

\subsection{Gradient Penalty}
\label{gradient_penalty}
As discussed in the introduction, without further regularisation deep networks
are prone to feature collapse. We find that it can be avoided by regularising
the representation map using a gradient penalty. Gradient penalties were first
introduced to aid generalisation in \citet{drucker1992improving}, who named it
``double backpropagation''. Recently, this type of penalty has been used
successfully in training Wasserstein GANs \citep{gulrajani2017improved} to
regularise the Lipschitz constant.

In our set up, we consider the following two-sided penalty:
\begin{align}
  \lambda \cdot \left[ ||\nabla_{x} \sum_c K_c ||_2^2 - 1 \right]^2,
\end{align}
where $||\cdot||_2$ is the $l_2$ norm and the targeted Lipschitz constant is 1.
We found empirically that regularising the gradient of $\sum_c K_c$ works better
than $f_\theta(x)$ or $\*K_c(x)$ (which is the vector of kernel distances for
input $x$). A similar approach was taken for softmax models by
\citet{ross2018improving}.

The two-sided penalty was introduced by \citet{gulrajani2017improved}, who
mention that despite a one-sided penalty being sufficient to satisfy their
requirements, the two sided penalty proved to be better in practice. The
one-sided penalty is defined as:
\begin{align}
  \lambda \cdot \max(0, ||\nabla_x \sum_c K_c ||_F^2 - 1).
\end{align}
In Section \ref{two_moons_exps}, we show the difference between the single and
two sided penalties experimentally. We find the two-sided penalty to be ideal
for enforcing sensitivity, while still allowing strong generalisation.

\subsection{Intuition about Gradient Penalty}

A gradient penalty enforces smoothness, limiting how quickly the output of a
function changes as the input x changes. Smoothness is important for
generalisation, especially if we are using a kernel which depends on distances
in the representation space. It is simple to show that regularising the $l_2$
norm of the Jacobian, $J$, enforces a Lipschitz constraint at least locally,
since for a small region around $x$ we have $g(x+ \epsilon) - g(x)\simeq J_g(x)
  \epsilon \le ||J(x)||_2 ||\epsilon||_2$.

However, smoothness still leaves us vulnerable to the feature collapse problem
outlined earlier, where multiple inputs are mapped to the same $g(x)$. Lipschitz
smooth functions can collapse their inputs --- the constant function $g(x) = c$
is Lipschitz for any Lipschitz constant $L$. Collapsing features can be
beneficial for accuracy, but it hurts our ability to perform out of distribution
detection, since it has the potential to make input points indistinguishable in
the representation space. We find empirically in our work that the two sided
penalty is extremely important: using the one sided penalty, i.e, enforcing only
smoothness, is not sufficient to produce the sensitive behaviour we want in our
representation. This can be seen in Figure \ref{two_moons_rbf_onesided}, in
contrast to Figure \ref{two_moons_duq} with the two-sided penalty.

By keeping the norm of the Jacobian above some value, intuitively we encourage
sensitivity of the learnt function, by preventing it from collapsing to a
locally constant function, ignoring all changes in the input space. This
argument is speculative, as this regularisation scheme has no effect on
sensitivity in directions orthogonal to the local Jacobian, and more work is
needed to explain definitively exactly why this penalty seems to encourage
sensitivity, as it would seem mathematically that collapsing the representation
would still be possible. However, we find empirically that it is important
for preserving out of distribution performance. In Appendix
\ref{gradient_penalty_analysis}, we evaluate a number of alternative approaches
such as using a reversible model as feature extractor (guaranteed to be
invertible) and computing the Jacobian with respect to the vector $K_c$ and
$f_\theta(x)$.

\subsection{Epistemic and Aleatoric Uncertainty}

When quantifying uncertainty, it can be useful to distinguish between
``epistemic'' and ``aleatoric'' uncertainty. Epistemic uncertainty comes from
uncertainty in the parameters of the model. This uncertainty is high for out of
distribution data, but also for example for informative data points in active
learning \citep{houlsby2011bayesian}. Aleatoric uncertainty is uncertainty
inherent in the data such as an image of a 3 that is similar to an 8
\citep{smith2018understanding}. In this case, the true class cannot be determined.
\begin{figure}[t]
  \includegraphics[width=0.9\linewidth]{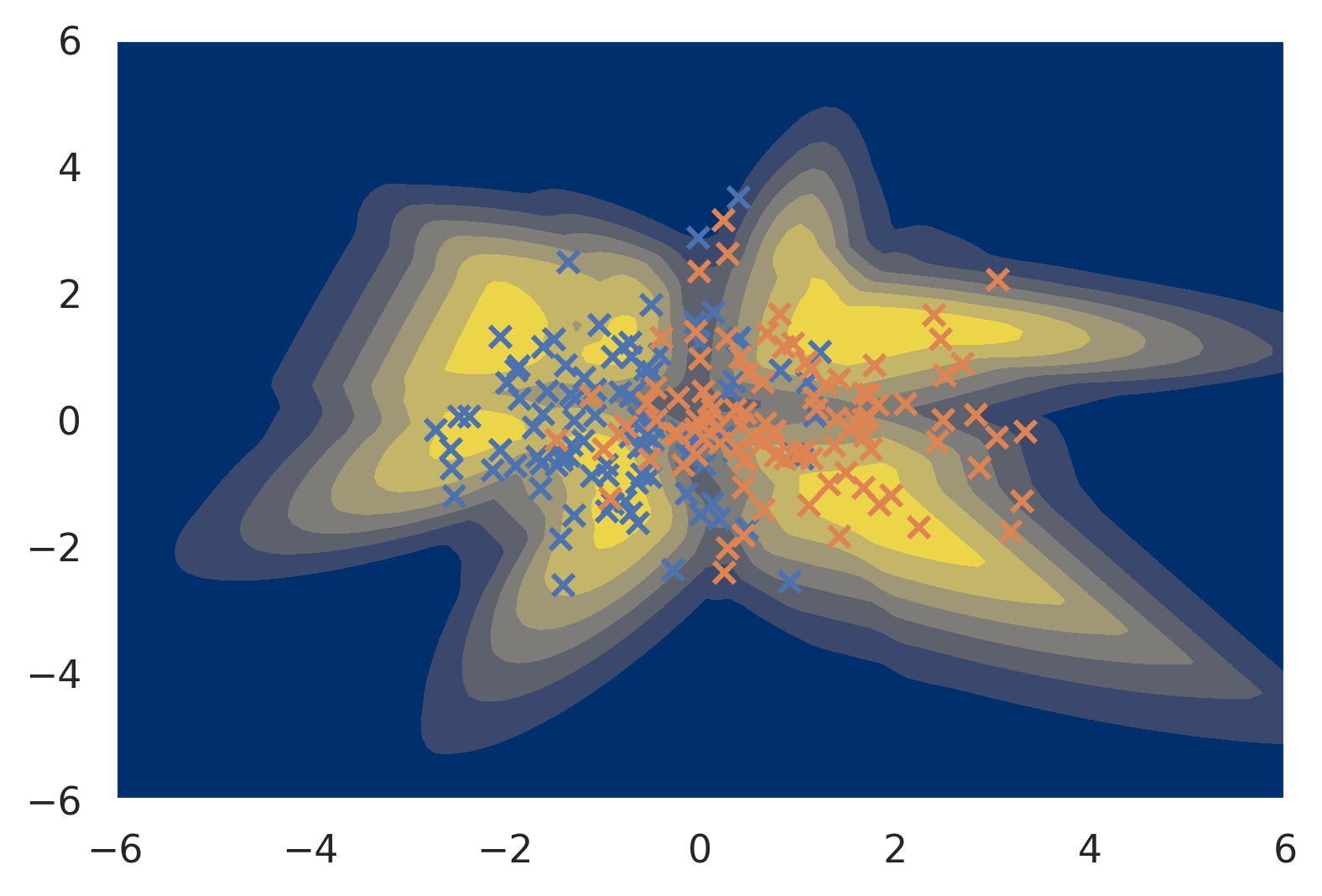}
  \caption{The uncertainty learned by DUQ on a simple problem of classifying
    samples from two overlapping Gaussian distributions. Yellow indicates
    certainty, while blue indicates uncertainty. There is significant aleatoric
    uncertainty due to the overlap between the classes. DUQ can express high
    aleatoric uncertainty by placing centroids close to each other in feature
    space, and is able to learn this in practice if the task needs it, as shown
    here by the higher uncertainty around the 0 mark on the x axis.}
  \label{aleatoric_uncertainty}
\end{figure}

In practice, DUQ captures both aleatoric and epistemic uncertainty. Informally,
when a point is far from all centroids in feature space there is epistemic
uncertainty. While aleatoric uncertainty is expressed by placing centroids so
that they are close in feature space (see Figure \ref{aleatoric_uncertainty})
and mapping a data point close to both of them. It is important that the
centroids are close in feature space, because otherwise the model would not map
them in between as it incurs a large loss, following Equation \ref{loss}. We do
not currently have a formal way to distinguish between these two kinds of
uncertainty in DUQ. Solving this problem is an interesting direction for future
research.

\subsection{Why Sensitivity can be at odds with Classification}
\label{accuracy_vs_sensitivity}

In this section we analyse some of the trade-offs and assumptions encoded in
detecting out-of-distribution inputs. We show in a toy experiment that standard
classification losses can hurt out-of-distribution detection. Consider fitting a
model on a problem with two features, $x_1$ and $x_2$, both sampled from a unit
Gaussian, and output y, such that $y = \operatorname{sign}(x_1) * \epsilon$,
where $\epsilon$ is noise with a low probability of flipping the label. The
optimal decision function in terms of the empirical risk, no matter the
algorithm, is the function $f(x_1, x_2) = \operatorname{sign}(x_1)$. But this
says nothing about the out of distribution behaviour. What happens if we now see
the input $x_1, x_2 = 1, 1000$? By our definition of the problem, this is out of
distribution, as it lies many standard deviations away from the observed data.
But should it be detected as out of distribution? The data does not define what
could be given as the input, at least if we take a conventional empirical risk
minimisation approach.

In this situation, it seems natural to prefer the kind of decisions which would
be made by a generative model, for example. If $x_1$ and $x_2$ represent medical
data, then presumably a highly abnormal value for $x_2$ is notable, and we would
like to detect it. However, if $x_2$ is a truly irrelevant variable, say, the
temperature on the surface of a distant planet, then presumably our model is
correct to ignore its value, even if the value of the irrelevant variable is
highly abnormal. When training using empirical risk minimisation, features not
relevant to classification accuracy can simply be ignored by the feature
extractors of a neural network. This makes out-of-distribution detection more
difficult using feature space methods, even those that use a distance loss as we
do. It is important to note that there is a potential tension here with
classification accuracy. Enforcing sensitivity can make accurate classification
harder because it forces the model to represent changes in input --- as in the
example above, these may be irrelevant to the causal structure of the problem.
If we know about invariances that are appropriate for the problem at hand, we
can enforce these by corresponding construction of the network. For example, we
enforce translation invariance by using convolutional networks in this paper.

\section{Related Work}

The largest body of research on obtaining uncertainty in deep learning are
Bayesian neural networks \citep{mackay1992bayesian, neal2012bayesian}. While
exact inference in them is intractable, a range of approximate methods have been
proposed. Mean-field variational inference methods, such as \textit{Bayes by
  Backprop} \citep{blundell2015weight} and Radial BNNs \citep{farquhar2020radial}
are a promising direction but have not yet lead to stable training on large
image datasets. A more scalable alternative is MC Dropout
\citep{gal2016dropout}, which is very simple to implement and evaluate. In
practice, these variational Bayesian methods are outperformed by Deep Ensembles
\citep{lakshminarayanan2017simple}. This is a simple, non-Bayesian, method that
involves training multiple deep models from different initialisations and a
different data set ordering. \citet{ovadia2019can} showed that Deep Ensembles
consistently outperform Bayesian neural networks that were trained using
variational inference. This performance comes at the expense of computational
cost, Deep Ensembles' memory and compute use scales linearly with the number of
ensemble elements at both train and test time.

Aside from using discriminative models, there have also been attempts at finding
out of distribution data using generative models. \citet{nalisnick2018deep}
showed that simply measuring the likelihood under the data distribution does not
work. Recently, a more advanced approach that involves separating the likelihood
of the semantic foreground from the background did show promising results on
selected datasets \citep{ren2019likelihood}. While generative models are a
promising avenue for out of distribution detection, they are not able to assess
predictive uncertainty; given that a data point is in distribution, can our
discriminative model actually make a reliable prediction? Further, generative
models are significantly more expensive to train than classification models.

Our approach is distinct from both ensembles/Monte Carlo methods, which aim to
find different explanations for the data and increase uncertainty when these
disagree, and generative models which model the data distribution directly.
Instead our approach is more related to pre-deep learning kernel methods
\citep{quinn2014least, scholkopf2000support}, such as Gaussian processes which
revert to a prior away from data, and Support Vector Machines, where the
distance to the separating hyperplane is informative of the uncertainty. These
approaches have never scaled to high dimensional data, because of a lack of well
performing kernel functions.

The decision function based on kernel distances was first used in the context of
convolutional neural networks by \citet{lecun1998gradient}. They were quickly
abandoned for softmax models, because they were difficult to scale and optimise
with gradient-based approaches due to saturating gradients and unstable
centroids. Notable improvements in our work over the original are the updating
mechanism of the centroids, solving the unstable centroids, and the loss
function that is based on a multivariate Bernoulli, solving saturating
gradients.

Regularising the Jacobian has a long history, starting with
\citet{drucker1992improving} and more recently \citet{ross2018improving}. Both
papers aim to regularise the $l_2$ norm of Jacobian down to zero. In the
first case to obtain better generalisation, while the second paper aims to
achieve adversarial robustness and interpretability. In neither case are the
authors interested in \textit{increasing} the Jacobian.
\citet{gulrajani2017improved} showed how a gradient penalty can be applied to
training GANs with the Wasserstein distance, which was a more scalable and
simpler alternative to weight clipping. They use the double sided penalty and
mention it works better in practice. Follow up work has analysed the penalty in
more detail and concluded that, contrary to our case, for training Wasserstein
GANS the one-sided penalty is preferable theoretically and practically
\citep{jolicoeur2019connections, petzka2017regularization}.

\section{Experiments}

We show the behaviour of DUQ in two dimensions, with the two moons dataset and
show the effect of leaving out the gradient penalty and using a one sided
penalty. We continue by looking at the out of distribution detection performance
for some notable difficult data set pairs \citep{nalisnick2018deep}, such as
FashionMNIST vs MNIST, and CIFAR-10 vs SVHN. We further study sensitivity to two
important hyper parameters the length scale $\sigma$ and gradient penalty weight
$\lambda$ and propose how to tune them without relying on example OoD data.

\subsection{Two Moons}
\label{two_moons_exps}
We use the scikit-learn \citep{pedregosa2011scikit} implementation of this
dataset and describe the model architecture and optimisation details in Appendix
\ref{two_moons_details}. For colouring the visualisations, we normalise the colour map
within the figure.

The result of our model trained with a two-sided gradient penalty is shown in
Figure \ref{two_moons_duq}. The uncertainty is exactly as one would expect for
the two moons dataset: certain on the training data, uncertain away from it and
in the heart within the two moons. The difference with Deep Ensembles is
striking (Figure \ref{two_moons_deep_ensembles}). The uncertainty for DUQ is
quantified as the distance to the closest centroid ($max$ over the kernel
distances), the uncertainty for Deep Ensembles is computed as the predictive
entropy of the average output, see Appendix \ref{deep_ensemble_uncertainty}. The
ensemble elements were trained separately using the same model as described in
Appendix \ref{two_moons_details}, but without L2 regularisation to encourage
diverse solutions.

\textbf{Discussion} While Figure \ref{two_moons_duq} is an impressive result in
deep learning, it is worth highlighting that Gaussian processes are able to
obtain such result too. A good visualisation can be found in
\citet{bradshaw2017adversarial}. Interestingly, even though Deep Ensembles have
been successfully applied to many large datasets \citep{ovadia2019can}, they
fail to estimate uncertainty well on the two moons dataset. This is due to the
simplicity and low dimensionality of this dataset, the ensembles generalise in
nearly the same way --- with a diagonal line dividing the top left and the
bottom right.
\begin{figure}[t]
  \begin{subfigure}{0.5\linewidth}
    \centering
    \includegraphics[width=\linewidth]{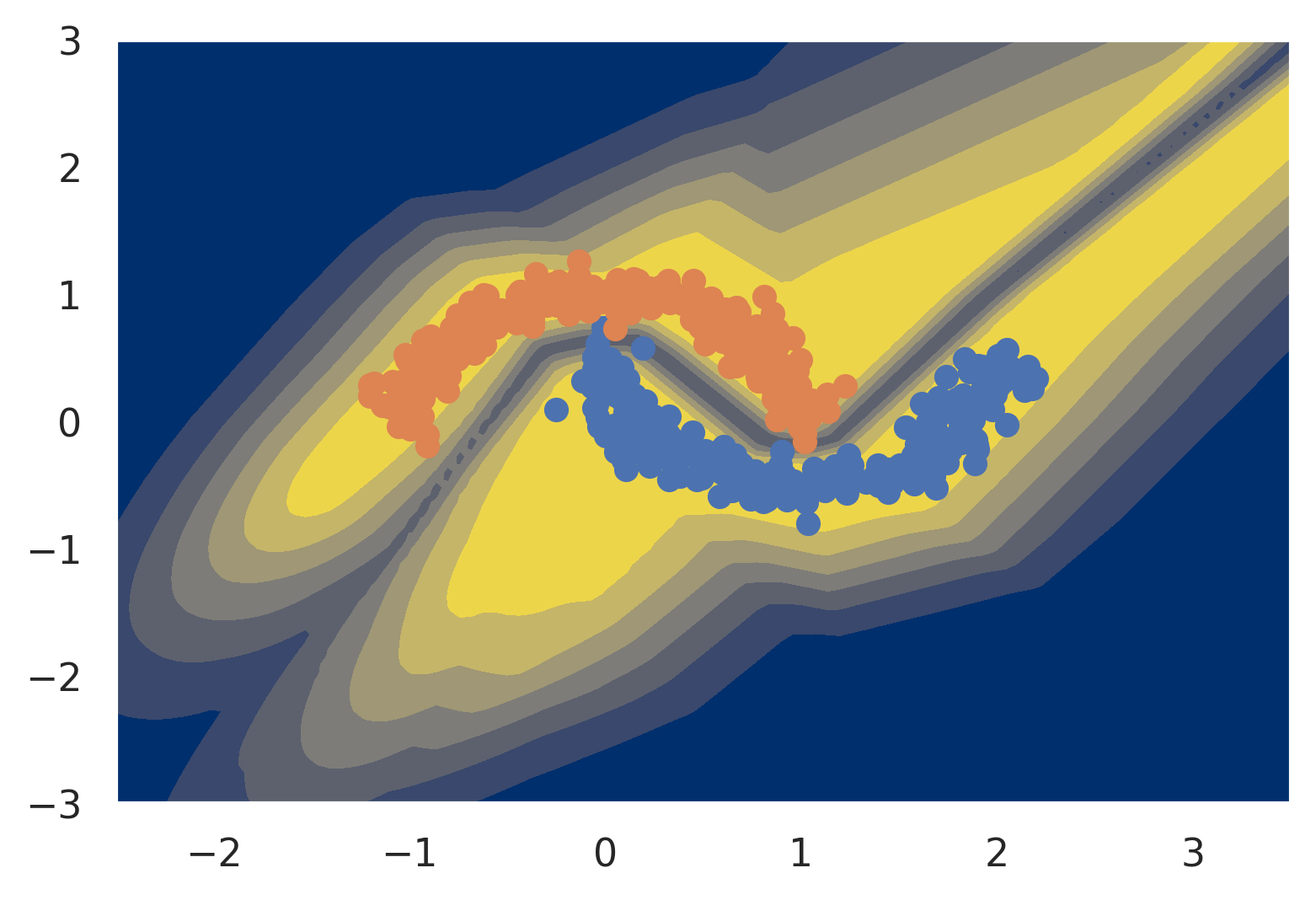}
    \caption{DUQ - No penalty}
    \label{two_moons_rbf}
  \end{subfigure}%
  \begin{subfigure}{0.5\linewidth}
    \centering
    \includegraphics[width=\linewidth]{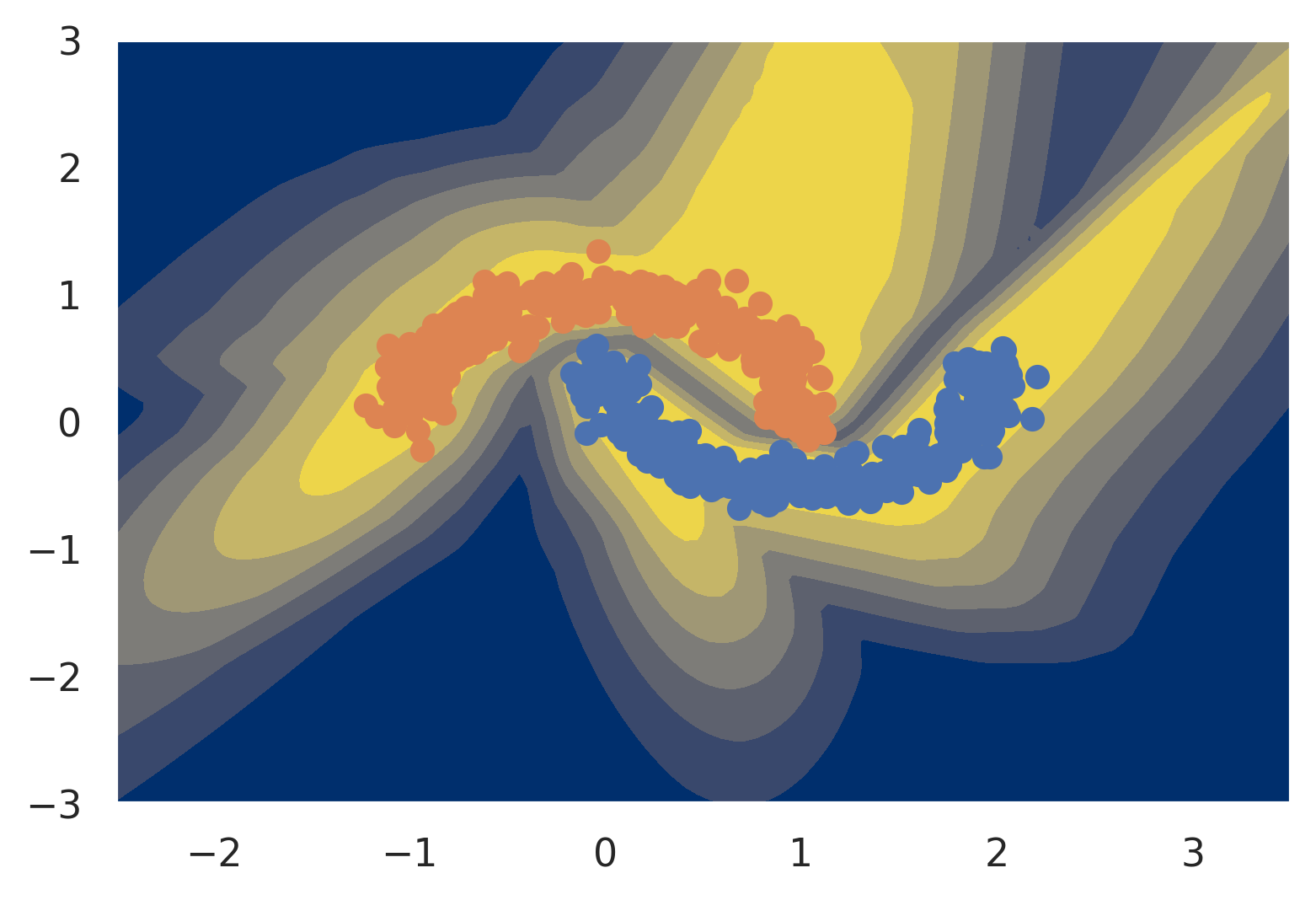}
    \caption{DUQ - One-sided penalty}
    \label{two_moons_rbf_onesided}
  \end{subfigure}
  \caption{Uncertainty results for two variations of DUQ: left without gradient
    penalty, and right with a one-sided gradient penalty ($\lambda = 1$). Yellow
    indicates certainty, while blue indicates uncertainty. Both results are
    significantly worse than DUQ with a two-sided penalty.}
  \label{two_moons_rbf_2}
\end{figure}

\textbf{Gradient Penalty} In Section \ref{gradient_penalty}, we introduced the
two-sided gradient penalty. Figure \ref{two_moons_rbf_2} shows why it is
important. In Figure \ref{two_moons_rbf}, we show the result of having no
gradient penalty, which shows that the model is certain every far away from the
data. In Figure \ref{two_moons_rbf_onesided}, we see that the uncertainty does
not improve when only a one-sided penalty is applied. In both cases, there are
'blobs' sticking out of the training data domain that are also classified with
high certainty.

\textbf{Hyper parameters} We found classification performance on two moons to be
insensitive to our setting of the gradient penalty weight $\lambda$, likely
because of the simplicity of the two moons dataset. For the uncertainty
visualisation, we found it important to set the length scale to be small (in the
interval $[0.05, 0.5]$), despite accuracy not being affected by this hyper
parameter. In the following experiments, we will discuss methods for picking the
length scale and the weight of the gradient penalty.

\subsection{FashionMNIST vs MNIST}

In this experiment, we assess the quality of our uncertainty
estimation by looking at how well we can separate the test set of FashionMNIST
\citep{xiao2017fashion} from the test set of MNIST \citep{lecun1998mnist} by
looking only at the uncertainty predicted by the model. We train our model on FashionMNIST and we
expect it to assign low uncertainty to the FashionMNIST test set, but high
uncertainty to MNIST, since the model has never seen that dataset before and it is
very different from FashionMNIST.

During evaluation we compute uncertainty scores on both test sets and measure
for a range of thresholds how well the two are separated. As in previous work
\citep{ren2019likelihood}, we report the AUROC metric, where a higher value is
better and $1$ indicates that all FashionMNIST data points have a higher
certainty than all MNIST data points. We picked FashionMNIST vs MNIST, because
it is a notably difficult dataset pair \citep{nalisnick2018deep}, while MNIST vs
NotMNIST \citep{bulatov2011notmnist} is much simpler.

\textbf{Experimental set up} Our model is a three layer convolutional network
and we report all architectural and optimisation details in Appendix
\ref{fashionmnist_details}. It is important to note that at test time we set
Batch Normalization to evaluation mode, meaning that we use the mean and
standard deviation of the feature activations computed from the training set
(i.e. FashionMNIST). It is unlikely that in practice we would get an entire
batch of (uncorrelated) OoD points, so we can not normalise using test time
batch statistics\footnote{Just one datapoint needs to have significantly
  different activation statistics for the entire batch to be easily detectable.}.
Further, we use the same data normalisation for the out of distribution set as
the in distribution set. Skipping either of these steps makes the problem
artificially simple.
\begin{figure}[t]
  \includegraphics[width=0.9\linewidth]{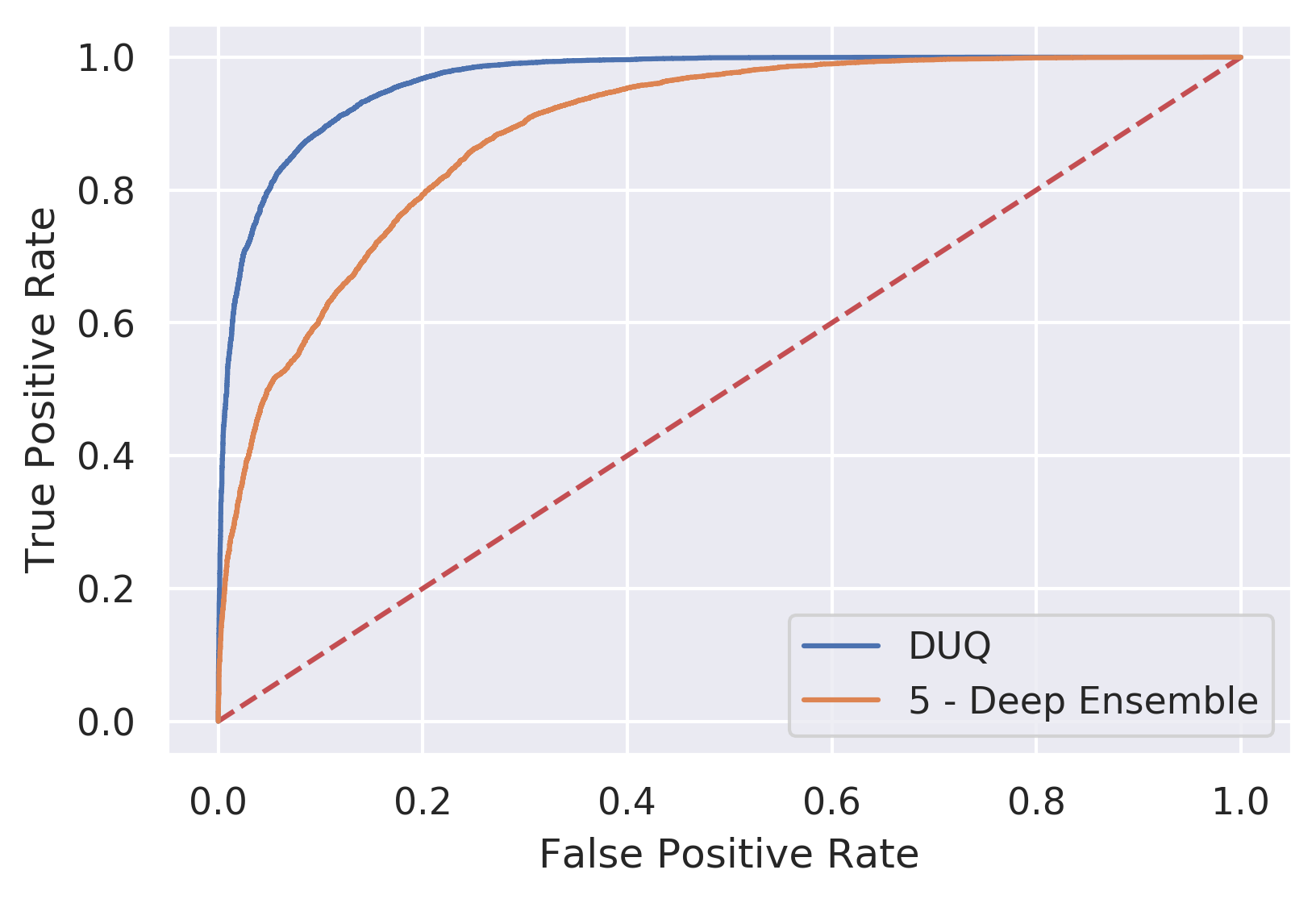}
  \caption{ROC curve for DUQ trained on FashionMNIST and evaluated on
    FashionMNIST and MNIST. The task is to separate these data sets based on
    uncertainty estimates.}
  \label{roc_curve_mnists}
  \vspace{-4mm}
\end{figure}

\textbf{Length Scale} Most hyper parameters, such as as the learning rate or
weight decay parameter, can be set using the standard train/validation split.
However there are two hyper parameters that are particularly important: the
length scale $\sigma$ and the gradient penalty weight $\lambda$. We set the
length scale by doing a grid search over the interval $(0, 1]$ while keeping
$\lambda = 0$. We pick the value that leads to the highest validation accuracy.
Following this process, we found that a length scale of $0.1$ leads to the
highest accuracy, as measured over five runs. While this process might not
result in a length scale that leads to the best OoD performance, it works well
in practice.

\textbf{Gradient Penalty} Setting the $\lambda$ parameter is more involved: from
Section \ref{accuracy_vs_sensitivity}, we know that the accuracy can suffer as a
result from gaining the ability to do out of distribution detection, so we
cannot rely on it to select the best $\lambda$. We also cannot use the AUROC
score on the MNIST dataset, because that would give the method an unfair
advantage: we cannot assume access to the OoD set in advance in
practice.\footnote{If we do assume access, then we can trivially train a binary
  classifier on the original and OoD set.} Instead we use a third dataset on which
we evaluate the AUROC and select our $\lambda$ values based on that. We follow
previous work \citep{ren2019likelihood} and use NotMNIST as the third dataset
for this pair. The results can be seen in Table \ref{lambda_results_table}. As
expected, the accuracy goes down as $\lambda$ increases, and we also observe
that the best AUROC result for NotMNIST coincides with the best score for MNIST,
which shows that the strategy of selecting a hyper parameter based on the
NotMNIST data set is reasonable. We note that while NotMNIST generalises to
MNIST, we cannot rely on this property in general. Therefore, we propose an
alternative method for model selection based on predictive uncertainty in
Section \ref{cifar10_results}.
\begin{figure}[t]
  \includegraphics[width=0.9\linewidth]{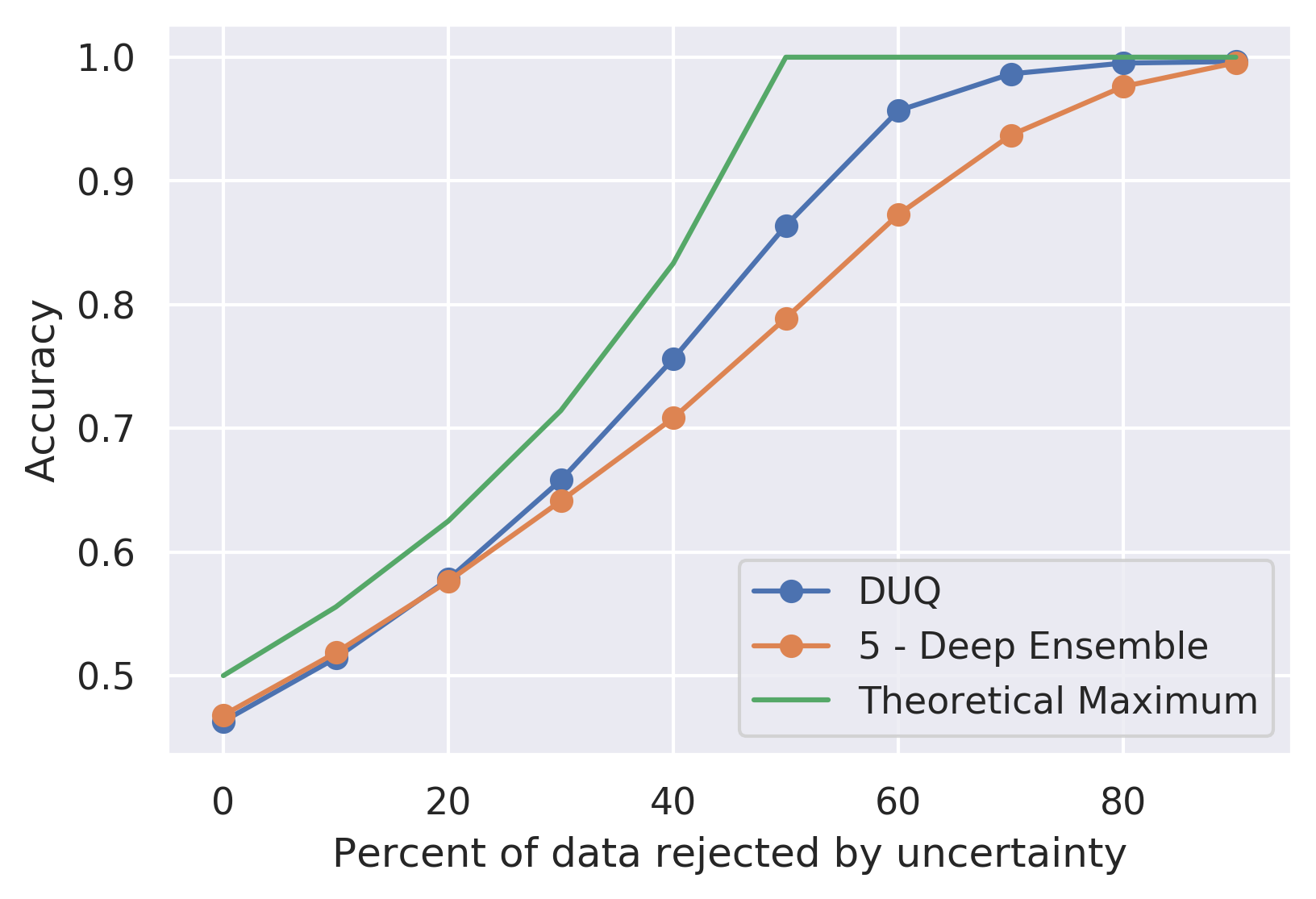}
  \caption{Rejection classification plot: accuracy on a combination of
    FashionMNIST and MNIST test sets. The x-axis indicates the proportion of data
    rejected based on the uncertainty score. The theoretical maximum is computed
    from a classifier with 100\% accuracy on FashionMNIST and rejects all MNIST
    points first.}
  \label{data_rejection_mnists}
\end{figure}

\textbf{Comparison} We show our results and compare with alternative methods in
Table \ref{fashionmnist_comparison}. Our proposed method, DUQ, outperforms all
other classification based methods. The only method that is better is LL ratio
\citep{ren2019likelihood}, which is based on generative models. These type of
models are more computationally costly to train than DUQ. The PixelCNN++
\citep{salimans2017pixelcnn} used by LL ratio for FashionMNIST uses 2 blocks of
5 gated ResNet layers, while our model is a simple three layer convolutional
network. An alternative, competitive approach is Mahalanobis Distance
\citep{lee2018simple}, which computes a distance in the feature space of a
pretrained softmax/cross entropy model in combination with a number of dataset
specific augmentations that rely on tuning via the out of distribution dataset
or in some cases an alternative third dataset.

The difference in AUROC between our Deep Ensemble result and
\citet{ren2019likelihood}'s is due to using different architectures. For a fair
comparison, we use the same architecture for the ensemble elements as for DUQ
(replacing the class dependent final layer by the usual single linear layer). In
Figure \ref{roc_curve_mnists}, we show the complete ROC curve for our
implementation of Deep Ensembles and DUQ. We see that DUQ outperforms Deep
Ensembles at all chosen rates.

\textbf{Accuracy and Gradient Penalty} To confirm that training using DUQ's
distance based output achieves competitive accuracy, we train two models using our
architecture: the standard softmax and cross entropy set up and DUQ with
$\lambda = 0$. We obtained $92.4\% \pm 0.1$ accuracy for the softmax model,
and for our proposed set up $92.4\% \pm 0.2$, both averaged over five runs.
The results show that we can obtain competitive accuracy using DUQ, resolving
previous problems with RBF networks. In Table \ref{lambda_results_table}, we
show how accuracy changes for an increasingly weighted gradient penalty. The
accuracy only degrades slightly, while AUROC is improved.
\begin{table}[b]
  \resizebox{\linewidth}{!}{
    \begin{tabular}{cccc}
      \toprule
      $\lambda$     & Acc (FM)                  & AUROC (NM)                 & AUROC (M)                  \\
      \hline
      0             & $92.4\% \pm .2$           & $0.933 \pm .009$           & $0.948 \pm .004$           \\
      \textbf{0.05} & $\mathbf{92.4\%  \pm .2}$ & $\mathbf{0.946  \pm .018}$ & $\mathbf{0.955  \pm .007}$ \\
      0.1           & $92.4\% \pm .1$           & $0.938 \pm .0018$          & $0.948 \pm .005$           \\
      0.2           & $92.2\% \pm .1$           & $0.945 \pm .019$           & $0.944 \pm .011$           \\
      0.3           & $92.3\% \pm .1$           & $0.944 \pm .013$           & $0.941 \pm .011$           \\
      0.5           & $92.0\% \pm .1$           & $0.946 \pm .014$           & $0.932 \pm .009$           \\
      1.0           & $91.9\% \pm .1$           & $0.945 \pm .018$           & $0.934 \pm .006$           \\
      \bottomrule
    \end{tabular}
  } \caption{FM stands for FashionMNIST, NM for NotMNIST, and M for MNIST. The
    results are mean/std computed from 5 experiment repetitions. We show AUROC for
    separating FashionMNIST from NotMNIST and MNIST; higher is better. We see that
    the gradient penalty improves AUROC performance slightly, but performance on
    this dataset pair is already very strong.}
  \label{lambda_results_table}
\end{table}

\textbf{Rejection Classification} In Figure \ref{data_rejection_mnists}, we
visualise how well these algorithms work in a more realistic scenario. We
combine the FashionMNIST and MNIST test sets, then we reject a certain portion
of the combined dataset by uncertainty score. Next we compute the accuracy on
the remaining data for each portion, considering all predictions on the OoD
MNIST set to be incorrect. We expect the accuracy to go up as we reject more of
the data points on which the model is uncertain. Ideally, we reject the
incorrectly classified FashionMNIST points and all MNIST points. The
\textit{Theoretical Maximum} is computed by assuming a model that has perfect
accuracy on the FashionMNIST test set \textbf{and} is able to reject all MNIST
data before any FashionMNIST data. This experiment combines out of distribution
detection, with detecting difficult to classify data points, which is closer to
actual deployment scenarios than the AUROC metric, and also a suggested
practically informed evaluation method by \citet{filos2019systematic}. Note that
the ensemble model has an accuracy of 93.6\% on FashionMNIST, giving it a 1.2\%
head start on DUQ, which has an accuracy of 92.4\%. We see that DUQ outperforms
Deep Ensembles in this more realistic scenario.
\begin{table}[b] \centering%
  \vspace{-4mm}
  \begin{tabular}{lc}
    \toprule
    Method                      & AUROC \\ \hline
    DUQ                         & 0.955 \\
    LL ratio (generative model) & 0.994 \\
    Single model                & 0.843 \\
    5 - Deep Ensembles (ours)   & 0.861 \\
    5 - Deep Ensembles (ll)     & 0.839 \\
    Mahalanobis Distance (ll)   & 0.942 \\
    \bottomrule
  \end{tabular}
  \caption{Results on FashionMNIST, with MNIST as OoD set. Deep Ensembles is by
    \citet{lakshminarayanan2017simple}, Mahalanobis Distance by
    \citet{lee2018simple}, LL ratio by \citet{ren2019likelihood}. Results marked by
    (ll) are obtained from \citet{ren2019likelihood}, (ours) is implemented using
    our architecture. Single model is our architecture, but trained with
    softmax/cross entropy.}
  \label{fashionmnist_comparison}
  \vspace{-4mm}
\end{table}
\begin{figure}[t]
  \centering
  \includegraphics[width=0.9\linewidth]{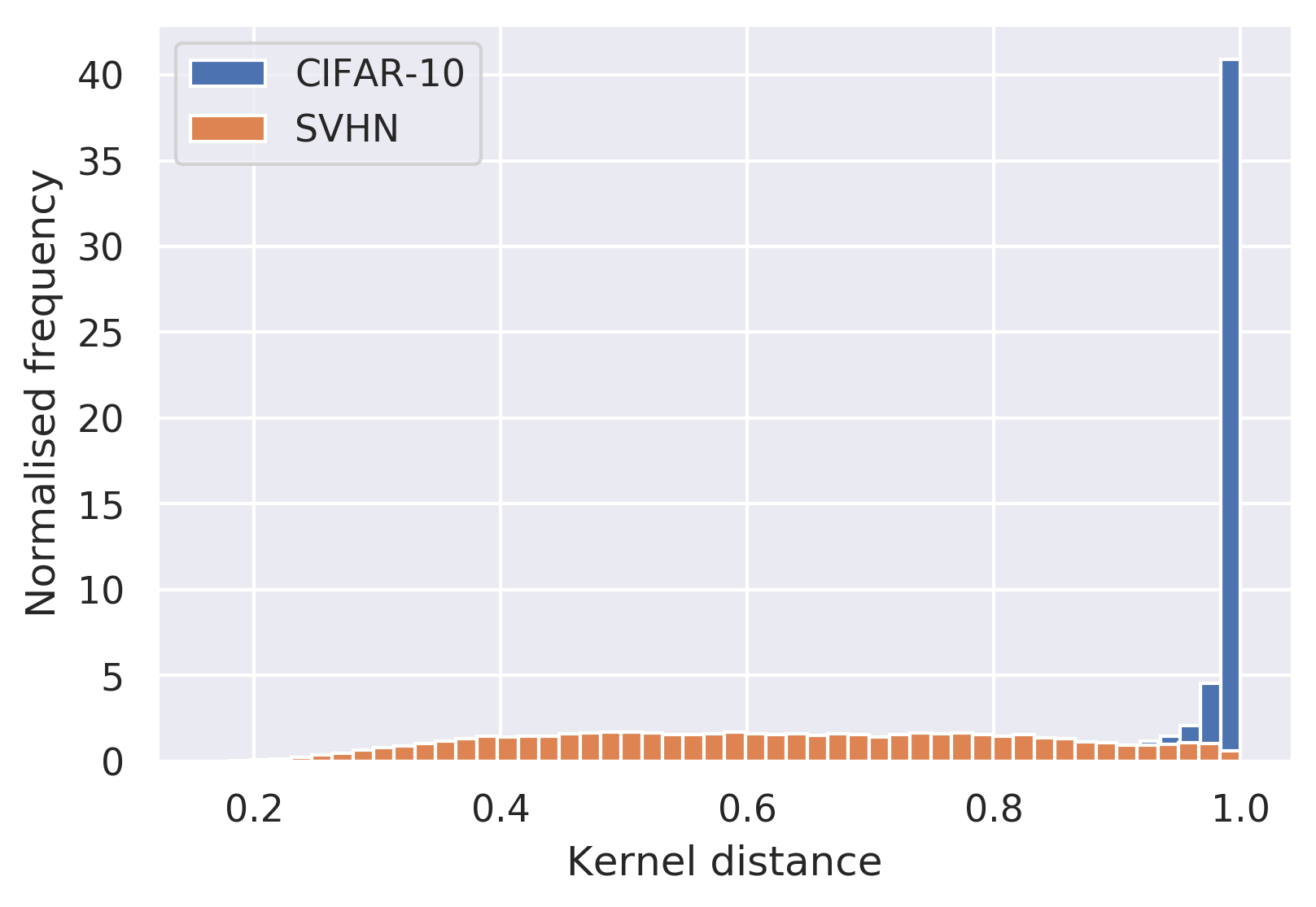}
  \caption{A histogram of uncertainty estimates as computed using DUQ ($\lambda
      = 0.5$). CIFAR-10 and SVHN are clearly separated. The counts are normalised,
    because the SVHN test set is significantly larger than CIFAR-10's.}
  \label{cifar_kernel_histogram}
\end{figure}
\subsection{CIFAR-10 vs SVHN}
\label{cifar10_results}
In this section we look at the CIFAR-10 dataset \citep{krizhevsky2014cifar},
with SVHN \citep{netzer2019street} as OoD set. We use a ResNet-18
\citep{he2016deep} as feature extractor $f_\theta(\cdot)$, specifically the
version provided by PyTorch \citep{paszke2017automatic} with some minor
modifications: we use 64 filters in the first convolutional layer, and skip the
first pooling operation and last linear layer. CIFAR-10 is a difficult dataset
for out of distribution detection for several reasons. There is a significant
amount of data noise: some of the dog and cat examples are not
distinguishable using only 32 by 32 pixels. The training set is small compared to
its complexity, making it easy to overfit without data augmentation.

\textbf{Experimental set up} As in the previous section, we tune the length
scale using the accuracy on the validation set, and find that $0.1$ works best
from a range of $[0.05, 1]$. We train for a fixed 75 epochs and reduce the
learning rate by a factor of $0.2$ at 25 and 50 epochs. We use random horizontal
flips and random crops as data augmentation and find that this is enough
regularisation to prevent the model from overfitting. All architectural and
optimisation details are described in Appendix \ref{cifar_details}. We obtain an
accuracy of $94.1\% \pm 0.2$ using the standard softmax/cross entropy loss. A
Deep Ensemble of several softmax models obtains an accuracy of $95.2\%$. DUQ
without a gradient penalty ($\lambda = 0$) obtains $94.2\% \pm 0.2$ accuracy,
while accuracy of DUQ with $\lambda = 0.5$ is $93.2\% \pm 0.4$.

\textbf{Gradient Penalty} For CIFAR-10, we do not use a third dataset to set
$\lambda$. Instead, we avoid using more data and use in-distribution
uncertainty. We measure this using the AUROC of detecting correctly and
incorrectly classified validation set data points using the predicted
uncertainty. We found that optimising $\lambda$ using this procedure also
transfer to $\lambda$ values that lead to strong out of distribution detection
performance. In general, this approach is preferable over using a third dataset,
because it is difficult to find an appropriate out of distribution dataset,
which will have the same characteristics as those encountered during deployment.
Imagine a particular difficult traffic situation or an MRI scan which shows a
new type of disease, these scenarios have no reasonable out of distribution set
available. Generative models are not able to take this approach, because they do
not have predictive uncertainty. Even if we use a hybrid model
\citep{nalisnick2019hybrid}, then the discriminative part, a softmax/cross
entropy model, does not have reliable predictive uncertainty.

\begin{figure}[t]
  \centering
  \includegraphics[width=0.9\linewidth]{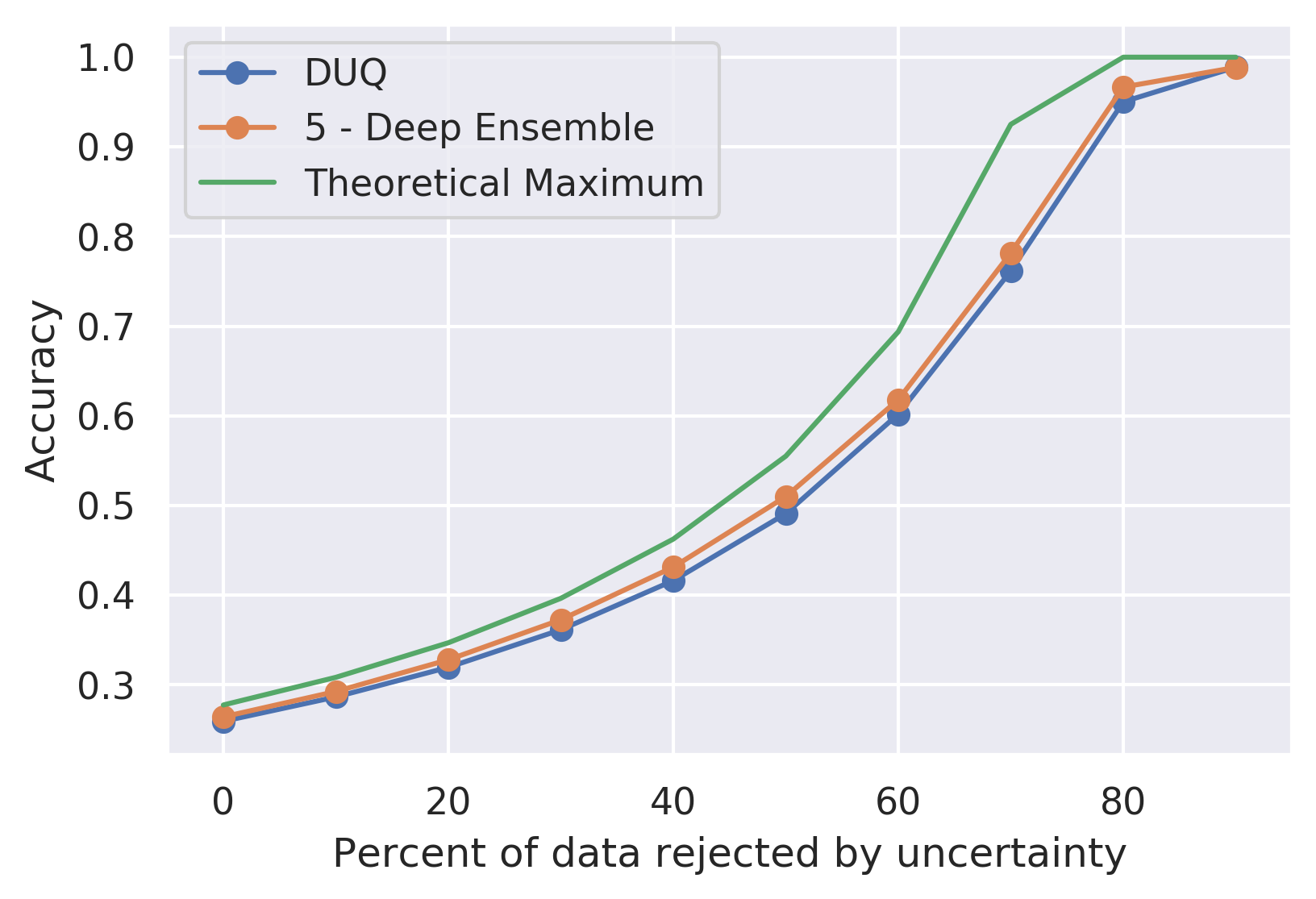}
  \caption{Rejection classification plot, which shows model performance on
    a mix of CIFAR-10 and SVHN, while rejecting uncertain points. The theoretical
    maximum is achieved when a hypothetical classifier obtains 100\% accuracy on
    CIFAR-10 and rejects all SVHN data points first. We see that DUQ and a 5 element
    Deep Ensemble perform very similar.}
  \label{cifar_data_rejection}
  \vspace{-4mm}
\end{figure}
\textbf{Results} In Figure \ref{cifar_kernel_histogram}, we show a a normalised
histogram for the kernel distances of CIFAR-10 and SVHN. We see that most of
CIFAR-10 is very close to 1, while SVHN is uniformly spread out over the range
of distances. This shows that DUQ works as expected and that out of distribution
data ends up away from all of the centroids in feature space.

The rejection classification plot, Figure \ref{cifar_data_rejection}, is created
similar to the previous experiment in the last section. Note that this time the
\textit{Theoretical Maximum} line is significantly lower, because the SVHN test
set contains close to $26,000$ elements, while CIFAR-10's only contains
$10,000$. This means that the best possible accuracy when 100\% of the data is
considered is about 28\%. We see that DUQ and Deep Ensembles perform similarly.

In Table \ref{cifar_comparison}, we compare DUQ with several alternative
methods. We see that DUQ performs competitively with a number of recent
approaches. Interestingly, on these more complicated data sets Deep Ensembles
performs the best. We suspect this is because the complexity of the data set
allows the ensemble elements to be more diverse while still explaining the data
well.

We further see a significant gap between DUQ with and without a gradient
penalty: there is a big improvement going from $\lambda = 0$ to $\lambda = 0.5$.
We suspect this is because there is a lot of within class variation, which
incentivises the model to collapse more diverse data points to the class centroids.

\textbf{Runtime} One of the main advantages of DUQ over Deep Ensembles is
computational cost. For Deep Ensembles, both computation and memory cost scale
linearly in the number of ensemble components, during both train and test time.
DUQ has to compute the Jacobian at training time, which is expensive, but at
test time there is only a marginal overhead over a softmax based model. Training
for one epoch on a modern 1080 Ti GPU, takes 21 seconds for a softmax/cross
entropy model, which leads to 105 seconds for a Deep Ensemble with 5 components.
DUQ with gradient penalty needs 103 seconds for one epoch at training time, but
only 27 seconds without gradient penalty. DUQ is ~25\% slower at test time than
single softmax/cross entropy model, but about 4 times faster than a Deep
Ensemble with 5 components.
\begin{table}[t] \centering%
  \begin{tabular}{lc}
    \toprule
    Method                      & AUROC             \\ \hline
    DUQ ($\lambda = 0.5$)       & $0.927 \pm 0.013$ \\
    DUQ ($\lambda = 0$)         & $0.861 \pm 0.032$ \\
    LL ratio (generative model) & 0.930             \\
    Single model                & $0.906 \pm 0.007$ \\
    3 - Deep Ensembles          & $0.926 \pm 0.010$ \\
    5 - Deep Ensembles          & $0.933 \pm 0.008$ \\
    10 - Deep Ensembles         & 0.941             \\
    15 - Deep Ensembles         & 0.942             \\
    \bottomrule
  \end{tabular}
  \caption{Deep Ensembles is by \citet{lakshminarayanan2017simple}, but
    re-implemented and evaluated using our architecture. LL ratio is as reported in
    \citet{ren2019likelihood}. Single model is our architecture, but trained with
    softmax/cross entropy. We show the AUROC for separating CIFAR-10 from SVHN.}
  \label{cifar_comparison}
  \vspace{-4mm}
\end{table}
\section{Conclusion}
We introduced DUQ, Deterministic Uncertainty Quantification, a simple method for
obtaining uncertainty using a deep neural network in a single forward
pass. Evaluations show that our method is better in some scenarios and
competitive in others with the more computationally expensive Deep Ensembles.

Interesting future work would be to place DUQ in a probabilistic framework, enabling
a calibrated notion of uncertainty and a rigorous way of separating out epistemic and
aleatoric uncertainty.

\section{Acknowledgements}
We thank Andreas Kirsch, Luisa Zintgraf, Bas Veeling, Milad Alizadeh, Christos
Louizos, and Bobby He for helpful discussions and feedback. We also thank the
rest of OATML for feedback at several stages of the project. JvA/LS are grateful
for funding by the EPSRC (grant reference EP/N509711/1 and EP/L015897/1,
respectively). JvA is also grateful for funding by Google-DeepMind.

\vfill\eject

\bibliography{duq}
\bibliographystyle{icml2020}
\clearpage

\appendix

\section{Experimental Details}

This section contains all info to reproduce our experiments. In all experiments, we
initialise the centroids using draws from $N(0, 0.05I_n)$. All convolutional layers are
initialized using the default in PyTorch 0.4.2 based on \citep{he2015delving}.

\subsection{Two Moons}
\label{two_moons_details}
We set the noise level to 0.1 and generate 1000 points for our training set. Our
model consists of three layers with 20 hidden units each, the embedding size is
10. We use the relu activation function and standard SGD optimiser with learning
rate 0.01, momentum 0.9 and L2 regularisation with weight $10^{-4}$. Our batch
size is 64 and we train for a set 30 epochs. We set the length scale to $0.3$,
$\gamma$ to 0.99 and $\lambda$ to 1.0.

\subsection{FashionMNIST}
\label{fashionmnist_details}
We use a model consisting of three convolutional layers of 64, 128 and 128 3x3
filters, with a fully connected layer of 256 hidden units on top. The embedding
size is 256. After every convolutional layer, we perform batch normalization and
a 2x2 max pooling operation.

We use the SGD optimizer with learning rate $0.05$ (decayed by a factor of 5
every 10 epochs), momentum $0.9$, weight decay $10^{-4}$ and train for a set 30
epochs. The centroid updates are done with $\gamma = 0.999$. The output
dimension of the model, $d$ is 256, and we use the same value for the size of
the centroids, $n$.

We normalise our data using per channel mean and standard deviation, as computed
on the training set. The validation set contains 5000 elements, removed at random
from the full 60,000 elements in the training set. For the final results, we rerun on
the full training set with the final set of hyper parameters.

\subsection{CIFAR-10}
\label{cifar_details}
We use a ResNet-18, as implement in torchvision version 0.4.2\footnote{Available
  online at: \url{https://github.com/pytorch/vision/tree/v0.4.2}}. We make the following
modifications: the first convolutional layer is changed to have 64 3x3 filters
with stride 1, the first pooling layer is skipped and the last linear layer is
changed to be 512 by 512.

We use the SGD optimizer with learning rate of $0.05$, decayed by a factor 10
every 25 epochs, momentum of $0.9$, weight decay $10^{-4}$ and we train for a
set 75 epochs. The centroid updates are done with $\gamma = 0.999$. The output
dimension of the model, $d$ is 512, and we use the same value for the size of
the centroids $n$.

We normalise our data using per channel mean and standard deviation, as computed
on the training set. We augment the data at training time using random
horizontal flips (with probability 0.5) and random crops after padding 4 zero
pixels on all sides. The validation set contains 10,000 elements, removed at
random from the full 50,000 elements in the training set. For the final results,
we rerun on the full training set with the final set of hyper parameters.

\section{Deep Ensemble Uncertainty}
\label{deep_ensemble_uncertainty}
The uncertainty in Deep Ensembles is measured as the entropy of the average
predictive distribution:

\begin{align*}
  \hat{p}(y|x)    & = \frac1N \sum_{i=1}^N p_{\theta_i}(y|x)             \\
  H(\hat{p}(y|x)) & = - \sum_{i=0}^C \hat{p}(y_i|x) \log \hat{p}(y_i|x),
\end{align*}

with $\theta_i$ the set of parameters for ensemble element $i$.
\begin{figure}[t]
  \centering
  \includegraphics[width=0.9\linewidth]{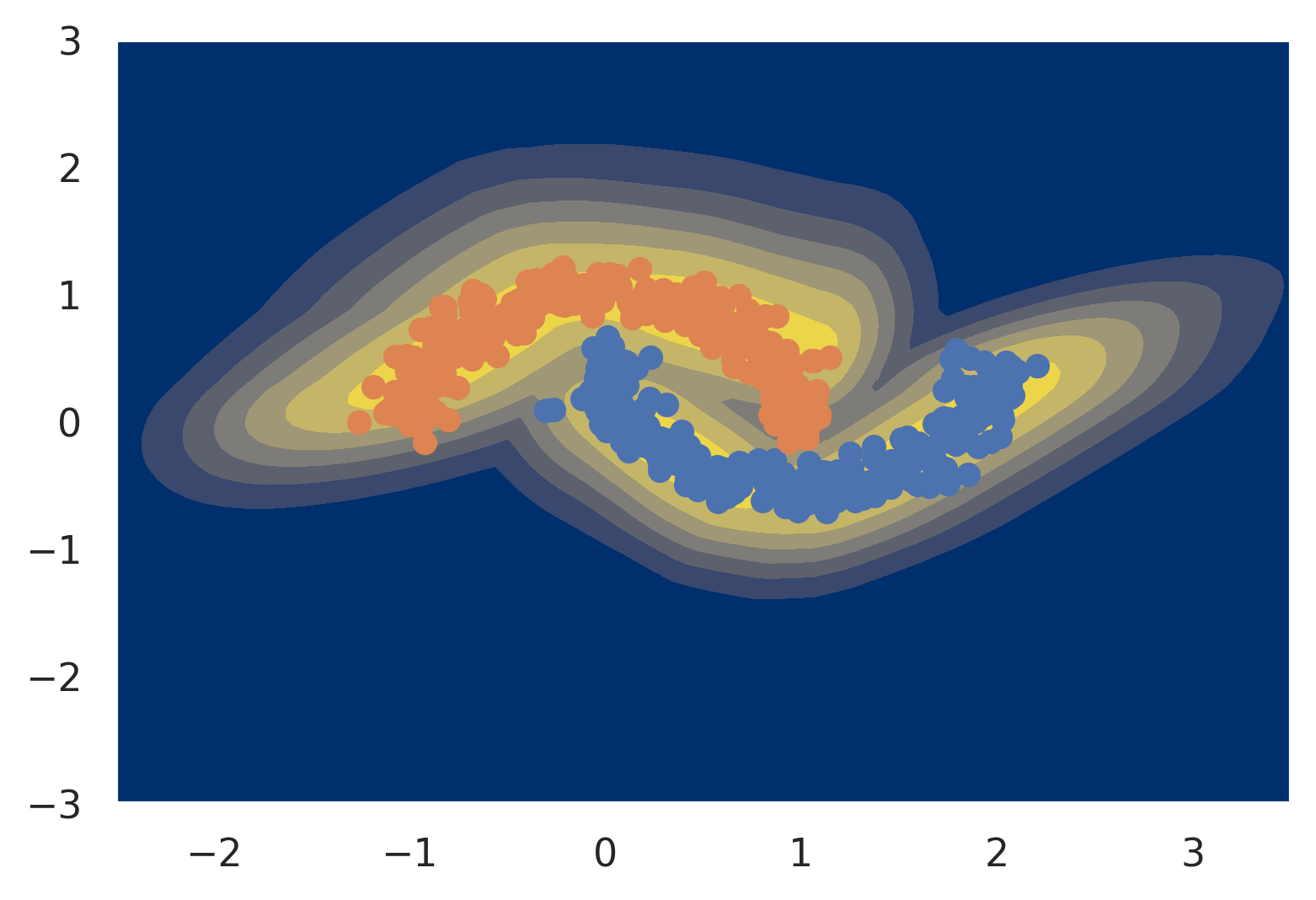}
  \caption{Uncertainty results on two moons data set. Yellow means certain,
    while blue indicates uncertainty. The model is a reversible feature extractor
    in combination with the kernel based output as in DUQ.}
  \label{two_moons_reversible}
\end{figure}

\section{Gradient Penalty Alternatives}
\label{gradient_penalty_analysis}

\subsection{Reversible Models}

An alternative method of enforcing sensitivity is by using an invertible feature
extractor. A simple and effective method of doing so is by using invertible
layers originally introduced in \citep{dinh2014nice}. Using these type of layers
leads to strong results on the two moons dataset as seen in Figure
\ref{two_moons_reversible}. Unfortunately, it is difficult to train reversible
models on higher dimensional data sets. Without dimensionality reduction, such
as max pooling, the memory usage of these type of networks is unreasonably high
\citep{jacobsen2018revnet} . We found it impossible to obtain strong accuracy
and uncertainty using these type of models, indicating that dimensionality
reduction is an important component of why these models work.

\subsection{Gradient Penalty}

Empirically, we found that computing the penalty on $\nabla_x \sum_c K_c$ works
well. However there are two other candidates to enforce the penalty on:
$\nabla_x K_c$, the vector of kernel distances and $\nabla_x f_\theta(x)$, the
feature vector output of the feature extractor. At first sight, these targets
might actually be preferential. Sensitivity of $f_\theta(x)$ ought to be
sufficient to obtain the out of distribution sensitivity properties we desire.

Computing the Jacobian of a vector valued output is expensive using automatic
differentiation. To evaluate the two alternative candidates we turn to the
Hutchinson's Estimator \citep{hutchinson1990stochastic}, which allows us to
estimate the trace of the Jacobian by computing the derivative of random
projections of the output. This approach was previously discussed in the context
of making neural networks more robust by \citet{hoffman2019robust}.

While we were able to get good uncertainty on the two moons data set using both
alternative targets, the results were not consistent. We attempted to reduce the
variance of the Hutchinson's estimator by using the same random projection for
each element in the batch, which worked well on two moons, but lead to
unsatisfactory results on larger scale data sets. In conclusion, we found that
while $\nabla_x \sum_c K_c$ is not a priori the best place to compute the
gradient penalty, it is still preferable over the noise that comes from applying
Hutchinson's estimator on any of the alternatives, at least in our experiments.

\end{document}